\documentclass[manuscript,screen]{acmart}

\usepackage{bm}
\usepackage{amsthm}
\usepackage{amsmath}
\usepackage{makecell}
\usepackage{subfigure}

\newtheorem{dfn}{Definition}[section]

\AtBeginDocument{%
  }

\begin{document}

\title{Learning with Noisy Ground Truth: From 2D Classification to 3D Reconstruction}







\author{Yangdi Lu}
\affiliation{%
  \institution{McMaster University}
  \city{Hamilton}
  \country{Canada}}
\email{luyangdi1992@gmail.com}

\author{Wenbo He}
\affiliation{%
  \institution{McMaster University}
  \city{Hamilton}
\country{Canada}}
\email{hew11@mcmaster.ca}

\renewcommand{\shortauthors}{Lu et al.}

\begin{abstract}
  Deep neural networks has been highly successful in data-intense computer vision applications, while such success relies heavily on the massive and clean data. In real-world scenarios, clean data sometimes is difficult to obtain. For example, in image classification and segmentation tasks, precise annotations of millions samples are generally very expensive and time-consuming. In 3D static scene reconstruction task, most NeRF related methods require the foundational assumption of the static scene (e.g.  consistent lighting condition and persistent object positions), which is often violated in real-world scenarios. To address these problem, learning with noisy ground truth (LNGT) has emerged as an effective learning method and shows great potential. In this short survey, we propose a formal definition unify the analysis of LNGT LNGT in the context of different machine learning tasks (classification and regression). Based on this definition, we propose a novel taxonomy to classify the existing work according to the error decomposition with the fundamental definition of machine learning. Further, we provide in-depth analysis on memorization effect and insightful discussion about potential future research opportunities from 2D classification to 3D reconstruction, in the hope of providing guidance to follow-up research.
\end{abstract}




\maketitle

\section{Introduction}
\label{sec:intro}

``Can machines think?'' This innovative question was raised in Alan Turing’s paper entitled ``Computing Machinery and Intelligence'' \citep{turing2009computing}. Suppose putting a machine player in an ``imitation game'', he stated that the best strategy for the machine is to try to provide answers that would naturally be given by a man. In other words, the ultimate goal of machines is to be as intelligent as humans. Over the past few decades, with the emergence of advanced models and algorithms (e.g. convolutional neural networks (CNNs) \citep{krizhevsky2012imagenet}, transformers \citep{vaswani2017attention}, GPTs \citep{radford2018improving}), large-scale data sets (e.g. ImageNet \citep{deng2009imagenet} with 1000 image classes), powerful computing frameworks and devices (e.g. GPU and distributed platforms), AI speeds up its pace to be like humans and supports many fields of daily life, such as search engines, autonomous driving cars, industrial robots and the recent popular chatGPT based on GPT-4 \citep{achiam2023gpt}.

Albeit its prosperity, the superior performance of current deep neural networks (DNNs) owns much to the availability of large-scale correctly annotated datasets, especially in supervised learning tasks. For example, image classification and segmentation always expect and assume a perfectly annotated large-scale training set. However, it is extremely time-consuming and expensive, sometimes even impossible to label a new dataset containing fully correct annotations. Typically, creating a regular dataset requires two steps: data collection and annotating process, involving two kinds of noise in the literature --- feature noise and ground truth noise \citep{zhu2004class}. Feature noise corresponds to the corruption of input data features, while ground truth noise refers to the change of ground truth from its actual one (e.g. in image classification task, by incorrectly annotating a dog label to a cat image). Both noise types can cause a significant decrease in the performance, while ground truth noise is considered to be more harmful \citep{frenay2013classification} as the ground truth is unique for each sample while features are multiple. For example, in video classification, video data contain audio, script and vision feature. The importance of each feature varies while the ground truth label always has a significant impact.

\begin{figure}[t]
	\begin{center}
		\includegraphics[width=0.7\linewidth]{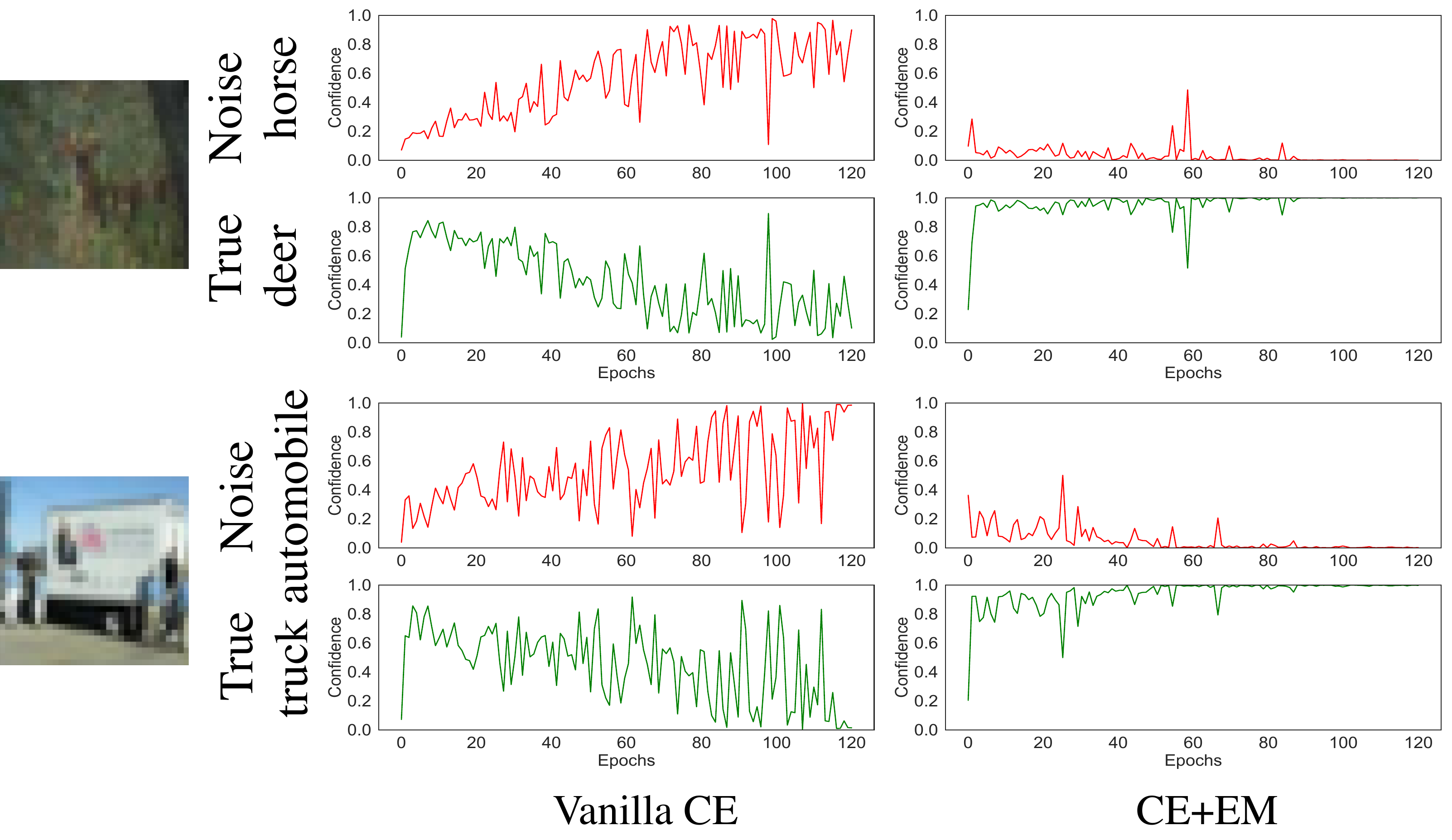}
	\end{center}
	\caption{\textbf{Effects of Noisy Labels.} Softmax outputs on noisy label and latent true label when training an 8-layer CNN on CIFAR10 with 40\% label noise. The x-axis is epochs and the y-axis is output probability on assigned label. We compare the Vanilla training using cross entropy (CE) loss with the method that adds entropy minimization (EM) of predictions to CE. The output probability of CE+EM is more stable than CE. }
	\label{fig:intro}
\end{figure}

To alleviate this problem, one may obtain the data with lower quality annotations efficiently through online keywords queries \citep{li2017webvision}. Similarly, the expensive annotating process can be crowdsourced with the help of platforms such as Amazon Mechanical Turk \footnote{http://www.mturk.com} and Crowdflower \footnote{http://crowdflower.com}, which effectively decrease cost. Another widely used approach is to annotate data with automated systems. However, all these approaches inevitably introduce the noisy ground truth. Moreover, noisy ground truth can occur even in the case of expert annotators, such as brain images \footnote{https://adni.loni.usc.edu/}. Even domain experts make mistakes because data can be extremely complex to be classified correctly \citep{lloyd2004observer}. Also, label noise can be injected intentionally to protect patients' privacy in image classification \citep{van2002randomized}.

In classification task, when DNNs model is trained with a noisy training set consisting of clean and mislabeled samples, it has been widely observed that the model outputs tend to severe fluctuate \citep{lu2022robust} then the model memorizes the noise. In Figure \ref{fig:intro}, we plot the softmax output probabilities corresponding to the noisy label and true label throughout the training. During the Vanilla training using cross entropy (CE) loss, the outputs can vibrate with large oscillations. Take the first row as an example, a deer image is mislabeled as a horse. In the training, the model begins with a high probability to indicate it is a deer image since the clean deer samples would encourage the model to predict this deer image as a deer. However, with the learning continuing, the deer samples with horse labels pull the model back to predict this deer image as a horse. Consequently, the model memorizes the wrong labels thus degrading the classifier prediction accuracy. In \citep{lu2022robust}, it was experimented that by simply adding a weighted entropy term to minimize prediction entropy constricts the randomness of model predictions, allowing the model to produce consistent and correct predictions. The right column in Figure \ref{fig:intro} shows the results after adding the entropy term to CE. It can be observed that the output probability on the latent true label become more stable compared to using CE.

In addition to the extensive use of DNNs in 2D classification tasks, the emerged 3D applications of novel view synthesis in fields such as robotics for action planning, 3D scene reconstruction, and AR/VR is gradually being explored with DNNs. Specifically, 3D scene reconstruction methods like Neural Radiance Fields (NeRF)~\cite{mildenhall2020nerf, barron2022mip} and 3D Gaussian Splatting (3DGS)~\cite{kerbl20233d} has greatly propelled the development of novel view synthesis. For example, NeRF-based methods have recently revolutionized 3D classical task, by storing 3D representations within the weights of a neural network. These representations are optimized by back-propagating image differences. When the fields store view-dependent radiance and volumetric rendering is employed we can capture 3D scenes with photo-realistic accuracy. However, these methods are built upon certain critical assumptions, including consistent lighting conditions and persistent object positions. These assumptions are frequently violated in real-world scenarios, which leads to artifacts, significant degradation in rendering quality. Thus, the recent popular challenge encountered in novel view synthesis is \emph{reconstructing the static and clean scene from noisy images containing distractors}, which disrupts the assumption of the static scene. In this survey, We show that it is similar to the learning with noisy labels in 2D classification and propose a simple idea to solve it.

Contributions of this survey can be summarized as follows:
\begin{itemize}
	\item We exploit and connect the \emph{memorization effect} of LNGT in 2D classification to 3D reconstruction. We are the first work to investigate the \emph{memorization effect} in 3D scene reconstruction (e.g .NeRF and 3DGS) optimization process.
	\item We provide a formal definition on LNGT, which naturally connects to the classic machine learning definition. The definition is not only general enough to include existing LNGT works but also specific enough to clarify what the goal of LNGT is and how we can solve it. 
	\item Based on our definition, we point out that the core issue of LNGT is the unreliable empirical risk minimizer, which is analyzed based on error decomposition in classic machine learning. This provides insights to improve LNGT in a more organized and systematic way and help us perform an extensive literature review.

	
\end{itemize}

\section{Memorization Effect}


\begin{figure}[t]
	\begin{center}
		\includegraphics[width=0.9\linewidth]{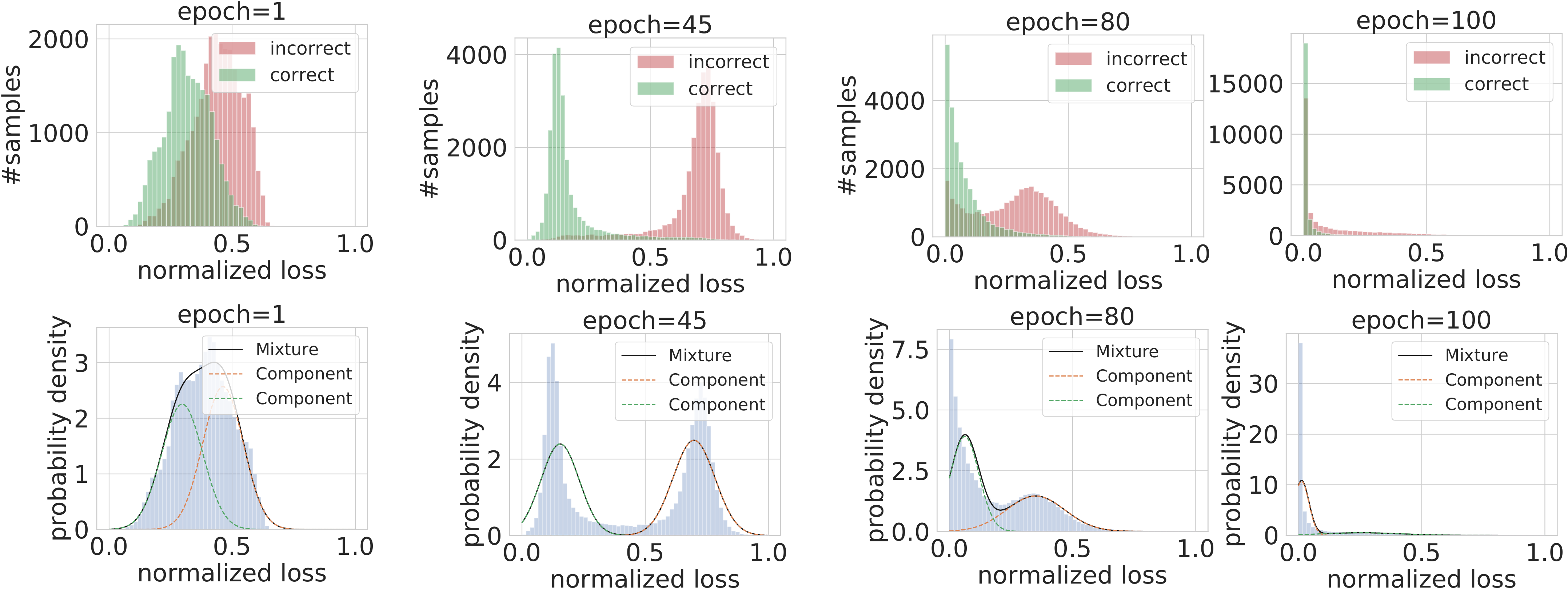}
	\end{center}
	\caption{\textbf{Memorization Effect in Image Classification.} We train ResNet34 on the CIFAR-10 with 60\% noise using CE loss and investigate the loss distribution. Top row: The normalized loss distribution over different training epochs. Bottom row: The corresponding mixture model after fitting a two-component GMM to loss distribution. Two components gradually separate at the beginning and start to merge with training continues.}
	\label{fig:class_dist}
\end{figure}

\subsection{2D Classification}

\textbf{Supervised Classification.} Considering a supervised classification problem with $K$ classes, suppose $\mathcal{X}\in\mathbb{R}^{d}$ be the input space, $\mathcal{Y}\in\{0,1\}^{K}$ is the ground-truth label space in an one-hot manner. In practice, the joint distribution $\mathcal{P}$ over $\mathcal{X}\times\mathcal{Y}$ is unknown. We have a training set $D=\{(\bm{x}_{i},\bm{y}_{i})\}^{N}_{i=1}$ which are independently sampled from joint distribution $\mathcal{P}$. Assume a mapping function class $\mathcal{F}$ wherein each $f:\mathcal{X}\rightarrow\mathbb{R}^{K}$ maps the input space to $K$-dimensional score space, we seek $f^{*}\in\mathcal{F}$ that minimizes an empirical risk $\frac{1}{N}\sum_{i=1}^{N}\ell(\bm{y}_{i},f(\bm{x}_{i}))$ for a certain loss function $\ell$.

\noindent\textbf{Classification with Noisy Labels.} Our goal is to learn from a noisy training distribution $\mathcal{P}_{\eta}$ where the labels are corrupted, with probability $\eta$, from their true distribution $\mathcal{P}$. Given a noisy training set $\bar{D}=\{(\bm{x}_{i},\bm{\bar{y}}_{i})\}^{N}_{i=1}$, the observable noisy label $\bm{\bar{y}}_{i}$ has a probability of $\eta$ to be incorrect. Suppose the mapping function $f$ is a deep neural network classifier parameterized by $\Theta$. $f$ maps an input $\bm{x}_{i}$ to $K$-dimensional logits $\bm{z}_{i}=f(\bm{x}_{i},\Theta)$. We obtain conditional probability of each class by using a softmax function $\mathcal{S}(\cdot)$, thus $\bm{p}_{i}=\mathcal{S}(\bm{z}_{i})$. Then the empirical risk on $\bar{D}$ using cross-entropy loss is
\begin{align}
	\label{eq:ce}
	\mathcal{L}_{\text{ce}}=\frac{1}{N}\sum_{i=1}^{N}\ell_{\text{ce}}(\bm{\bar{y}}_{i},\bm{p_{i}})=-\frac{1}{N}\sum_{i=1}^{N}(\bm{\bar{y}}_{i})^{\top}\log(\bm{p}_{i}).
\end{align} 
When directly optimizing $\mathcal{L}_{\text{ce}}$ by stochastic gradient descent (SGD), the DNNs have been observed to completely fit the training set including mislabeled samples eventually (see Fig. \ref{fig:class_dist} right most column), resulting in the test performance degradation in the later stage of training. In addition, the clean samples tend to have smaller loss values than the mislabeled samples in early stage \citep{lu2022noise}. We analyze the normalized loss distribution over different training epochs in Fig. \ref{fig:class_dist} top row. Intriguingly, the two distributions are merged at the initialization, then start to separate, but resume merging after the certain point. 

To alleviate the impact of noisy labels in training data, existing work Bootstrap \cite{reed2014training} proposes to generate soft target by interpolating between the original noisy distributions and model predictions by $\beta\bm{\bar{y}}+(1-\beta)\bm{p}$, where $\beta$ weights the degree of interpolation. Thus the cross-entropy loss using Bootstrap becomes
\begin{align}
	\label{eq:bs}
	\mathcal{L}_{\text{bs}}=-\frac{1}{N}\sum_{i=1}^{N}\Big(\beta\bm{\bar{y}}_{i}+(1-\beta)\bm{p}_{i}\Big)^{\top}\log(\bm{p}_{i}).
\end{align}
A static weight (e.g. $\beta=0.8$) is applied as an approximate measure for the correction of a hypothetical noisy label. Another work M-correction \cite{arazo2019unsupervised} makes $\beta$ dynamic for different samples, i.e., using a noise model to individually weight each sample. 
\begin{align}
	\label{eq:mc}
	\mathcal{L}_{\text{mc}}=-\frac{1}{N}\sum_{i=1}^{N}\Big(w_{i}\bm{\bar{y}}_{i}+(1-w_{i})\bm{p}_{i}\Big)^{\top}\log(\bm{p}_{i}).
\end{align}
$w_{i}$ is dynamically set to posterior probability conditioned on loss value and the Gaussian Mixture Model (GMM) is estimated after each training epoch using the normalized cross entropy loss for each sample $i$. Thus, correct samples rely on their given label $\bar{y}_{i}$ ($w_{i}$ is large), while incorrect ones let their loss being dominated by their class prediction $z_{i}$ or their predicted probabilities $p_{i}$ (1 - $w_{i}$ is large). In early mature stages of training the CNN model should provide a good estimation of the true class for noisy sample, shown in Fig \ref{fig:class_dist} when epoch = 45.

\begin{figure}[t]
	\centering
	\subfigure[Mip-NeRF 360 \cite{barron2022mip}.]{\label{fig:1}
		\includegraphics[width=0.8\columnwidth]{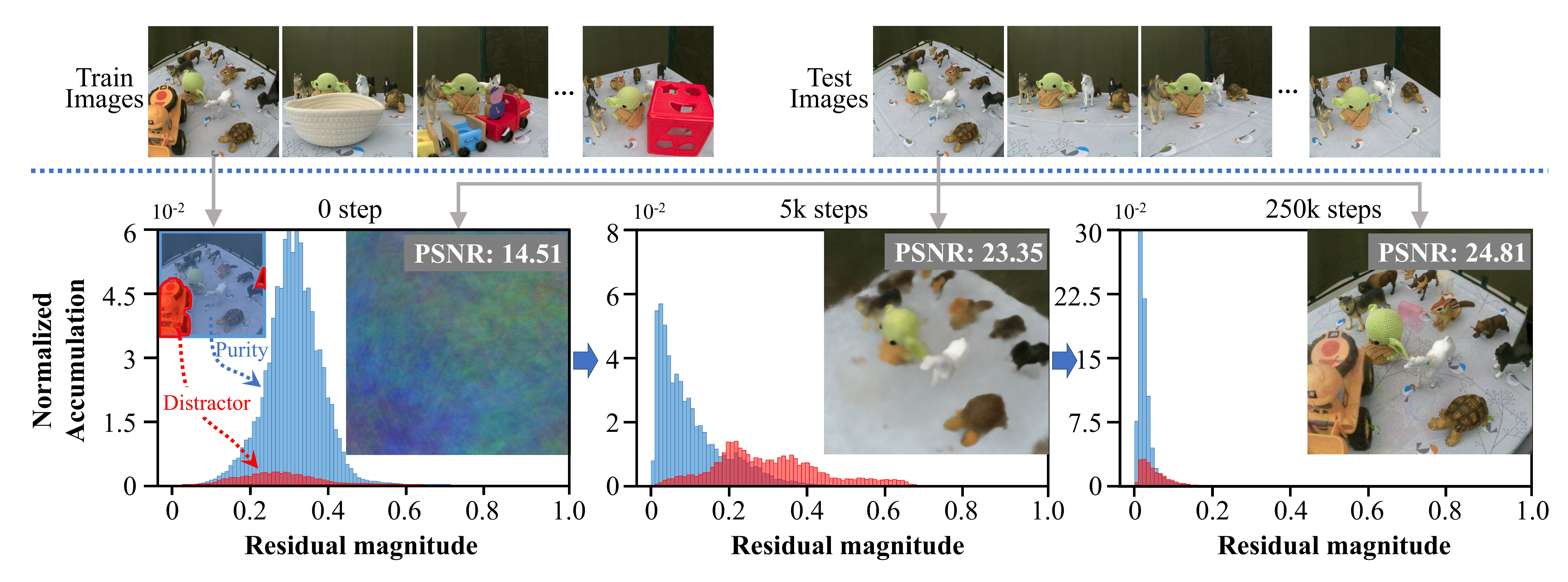}}
	\subfigure[Mip-NeRF 360 with dynamic weight mask $\mathbf{M}_{r}$ estimated by GMM.]{\label{fig:2}
		\includegraphics[width=0.8\columnwidth]{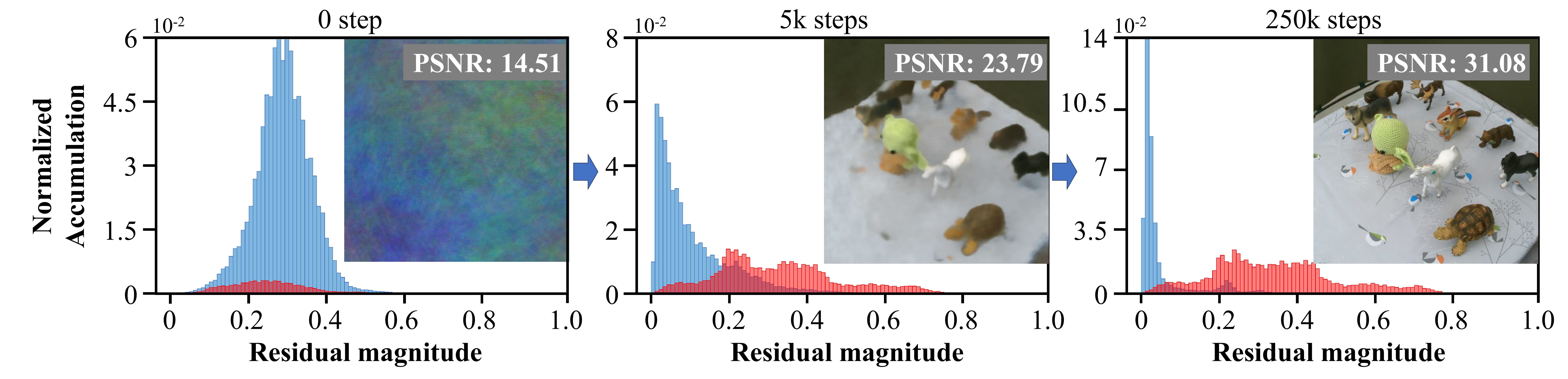}}
	\caption{\textbf{Memorization Effect in 3D reconstruction.} {The purity (clean scene) and distractor pixels correspond to the blue and red bars in the histograms, respectively. We observe that during the initial stages of optimization (e.g., the 5000th step), the image exhibits blurring, yet the scene remains relatively clean. However, as optimization progresses, the image sharpens, concomitant with the emergence of distractors. Further analyzing the accumulated loss distribution, Mip-NeRF 360 primarily focuses on learning purity pixels (of the clean scenes) and leaves most distractor pixels out of the learning process at the early stage, as evidenced by the minimal changes in the histogram of distractor pixels (red bars).}}
	\label{fig:nerf}
\end{figure}

\subsection{3D Scene Reconstruction} 
Neural Radiance Fields (NeRF)~\cite{mildenhall2020nerf} has made a breakthrough in novel view synthesis, which is capable of synthesizing photo-realistic images at arbitrary views. NeRF represents a 3D scene as a continuous radiance field, and synthesizes images with differentiable volumetric rendering. Recently, 3D Gaussian Splatting~\cite{kerbl20233d} has emerged as another promising paradigm for novel view synthesis, which synthesizes the image by rendering a set of learnable Gaussian points. Despite great development in novel view synthesis, the aforementioned methods are limited to static scenes with constant light conditions. They are struggling with multi-view images that contain distractors. 

Existing novel view synthesis methods represent a 3D scene as a parameterized model $\mathbf{S}_\theta$, while facilitating a differentiable rendering technique $\mathcal{R}(\cdot,\cdot)$ to render an image $\hat{\mathbf{I}} \in \mathbb{R}^{{H}\times {W}}$ as following:
\begin{equation}
	\hat{\mathbf{I}} = \mathcal{R}(\mathbf{S}_\theta,\pi),
\end{equation}
where $\pi$ is the image pose. Concretely, Neural Radiance Fields~\cite{mildenhall2020nerf} leverages the Multilayer Perceptron (MLP) and volumetric rendering, whereas 3D Gaussian Splatting~\cite{kerbl20233d} is built upon a collection of learnable Gaussian points and splatting rasterization. However, both of them optimize a $\mathcal{L}_{2}$ loss between the ground truth image and rendered image:
\begin{equation}
	\mathcal{L}_{nerf} = \sum_{r} ||\hat{\mathbf{I}}_r - \mathbf{I}_r||_2^2,
\end{equation}
where $\hat{\mathbf{I}}_r$ and $\mathbf{I}_r$ are the rendered color and ground truth color at pixel $r$, respectively. In the training of NeRF and 3DGS with multi-view images containing distractors (noise), we observed the similar phenomenon of \emph{memorization effect} as in classification task. This phenomenon refers to the parameters initially fitting purity and excluding distractor pixels in the early stages of training, gradually overfitting to distractor pixels in the later stages, as evidenced by the rendered image and histogram in Fig. \ref{fig:nerf} (a). It degrades the quality of rendered image in novel views. Most of the 3D reconstruction methods do not consider the case of distractor noise, while the most related work is RoubstNeRF~\cite{sabour2023robustnerf}.
\begin{figure}[t]
	\begin{center}
		\includegraphics[width=0.95\linewidth]{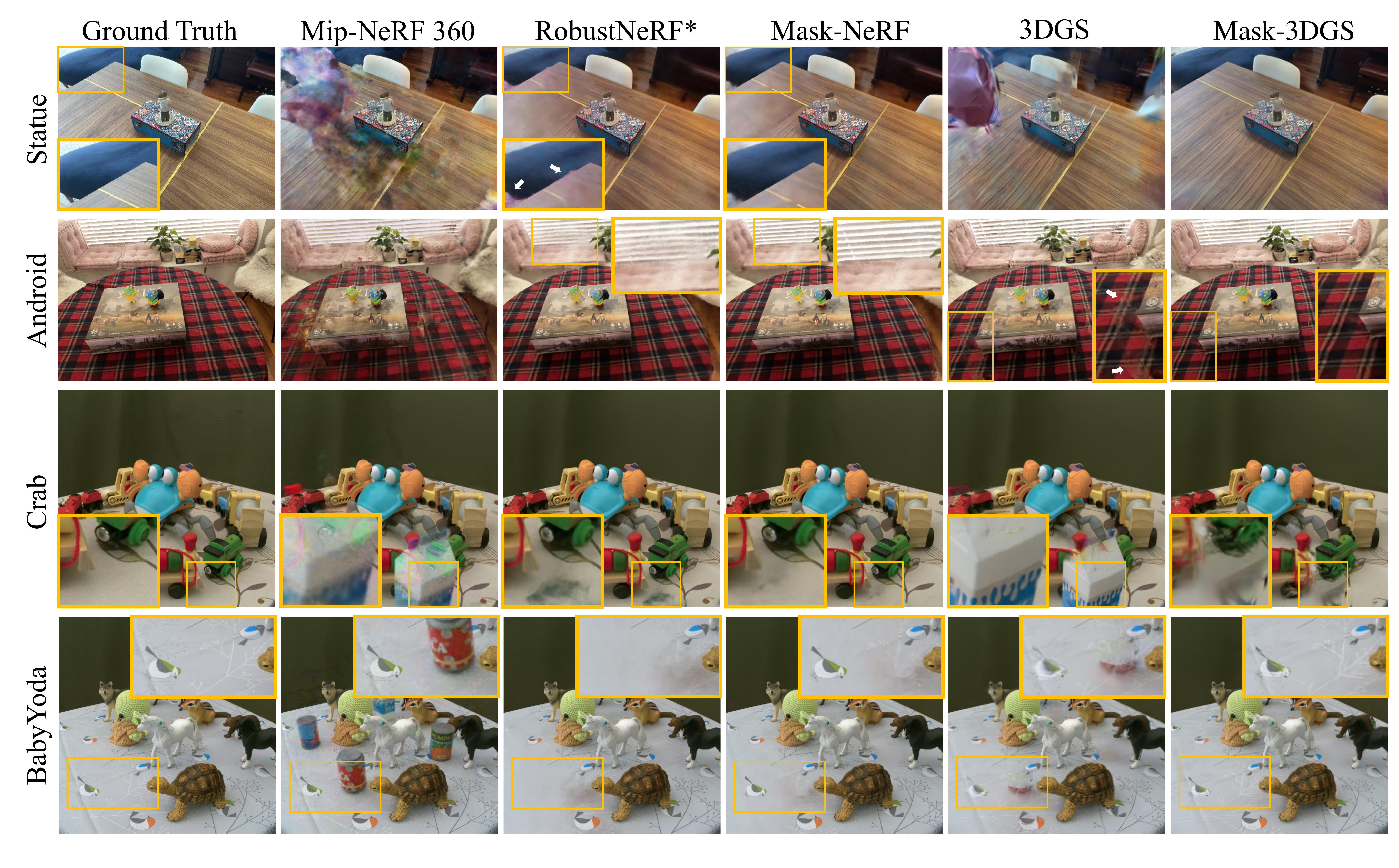}
	\end{center}
	\caption{\textbf{Qualitative Comparison on the RobustNeRF Dataset.} Compared to the baselines,  Mip-NeRF 360 and 3DGS, our Mask-NeRF and Mask-3DGS not only efficiently eliminate distractors but also retain a higher level of detail. In comparison with RobustNeRF*, MemE models demonstrate superior performance in removing distractors, leading to a reduction in artifacts and an enhancement in detail preservation.}
	\label{fig:yoda_re}
\end{figure}
To improve it, we can simply generate a dynamic weight $\mathbf{M}_{r}$ to indicate the distractor pixels similar as Equation \eqref{eq:mc}. Then we analogy noisy ground truth color $\mathbf{I}_r$ to noisy ground truth label $\bar{y}_{i}$ and rendered color $\hat{\mathbf{I}}_r$ to predicted probabilities $p_{i}$. Then the loss of new mask-NeRF is
\begin{align}
	\label{eq:m-nerf}
	\mathcal{L}_{mask-nerf} = \sum_{r} || \big(\mathbf{M}_r  \cdot \hat{\mathbf{I}}_r + (1 - \mathbf{M}_r) \cdot \mathbf{I}_r \big)- \mathbf{I}_r||_2^2= \sum_{r} || \mathbf{M}_r  \cdot \hat{\mathbf{I}}_r + \mathbf{I}_r - \mathbf{M}_r \cdot \mathbf{I}_r - \mathbf{I}_r||_2^2 = \sum_{r}\mathbf{M}_r ||\hat{\mathbf{I}}_r - \mathbf{I}_r||_2^2
\end{align}
Similar to the M-correction in classification task, leverages the \emph{memorization effect} to effectively prevent distractors from overfitted, maintaining the position of the red bars unchanged as much as possible. Similar to classification, early mature stage (5k steps) enables a clear distinction between purity and distractor pixels. Subsequently, we can promote the fitting of purity pixels and suppress the learning of distractor pixels throughout the Mip-NeRF 360 training process to easily achieve sharp and clean view rendering in a noisy input scenario. For estimation of $\mathbf{M}_r$, we can use the GMM or other unsupervised model to fit the loss distribution to differentiate the clean pixels from distractor (noisy) pixels. One case results (BabyYoda) with GMM on RoubustNeRF dataset~\cite{sabour2023robustnerf} are shown in Fig. \ref{fig:nerf} (b). More qualitative comparison results with existing methods are shown in Fig. \ref{fig:yoda_re}. This research direction is barely explored in 3D reconstruction area, more work need to be done to make the algorithm more robust in the near future.





\section{Formal Definition of LNGT}
LNGT is a sub-area in machine learning, before giving the formal definition of LNGT, let us recall how machine learning is defined in the existing literature.
\begin{dfn}[Machine Learning \citep{mitchell1997machine}]
	\label{def:1}
	A computer program is said to \textbf{learn} from experience $E$ with respect to some classes of task $T$ and performance measure $P$, if its performance at tasks in $T$, as measured by $P$, can improve with $E$.
\end{dfn}
The above definition can be generalized to a very wide range of practical applications. For example, consider an image classification task $(T)$, a machine learning program improves its classification accuracy $(P)$ through $E$ obtained by training on a large number of labeled images (e.g. the ImageNet). Typically, existing machine learning applications, especially using deep neural networks as in the example mentioned above, require a lot of data samples with correct supervision information. However, this may be difficult or sometimes even impossible in real-world applications. LNGT is a special and more general case of machine learning, which targets at obtaining good learning performance given noisy supervised information in the training set, which consists of examples of inputs $\bm{x}_{i}$'s along with their corresponding output $\bar{y}_{i}$'s. Formally, we define LNGT in Definition \ref{def:2}.

\begin{dfn}
	\label{def:2}
	\textbf{Learning with Noisy Ground Truth} (LNGT) is a type of machine learning problems (specified by $\bar{E}$, $T$ and $P$), where $\bar{E}$ is corrupted version of invisible clean $E$, consists of clearn and wrong examples for the target task $T$.
\end{dfn}

In any machine learning problem, usually there are prediction errors and one cannot obtain perfect predictions. In this section, we illustrate the core issue of LNGT based on error decomposition in supervised machine learning \citep{bottou2007tradeoffs}. This analysis applies to LNGT including classification and regression tasks.

\subsection{Notations}
Consider a learning task $T$,  LNGT deals with a dataset $\bar{D}=\{\bar{D}_{\text{train}},D_{\text{test}}\}$ consisting of a noisy training set $\bar{D}_{\text{train}}=\{(\bm{x}_{i},\bar{y}_{i})\}^{N}_{i}$, and a clean testing set $D_{\text{test}}$. Let $p(\bm{x},y)$ be the ground truth joint probability distribution of input $\bm{x}$ and output $y$, $\bar{p}(\bm{x},\bar{y})$ be the corrupted joint probability distribution of input $\bm{x}$ and output $\bar{y}$. Let $\hat{h}$ be the optimal hypothesis from $\bm{x}$ to $y$. LNGT learns to discover $\hat{h}$ by fitting $\bar{D}_{\text{train}}$ and testing on $D_{\text{test}}$. For clarity, we assume $\bar{D}_{\text{c\_train}}$ be a set of clean training samples (i.e. inputs with correct labels) and $\bar{D}_{\text{m\_train}}$ be the mislabeled training samples (i.e. inputs with wrong labels). We have $\bar{D}_{\text{train}}=\bar{D}_{\text{c\_train}} \cup \bar{D}_{\text{m\_train}}$. Note that $\bar{D}_{\text{c\_train}}$ and $\bar{D}_{\text{m\_train}}$ are imagination sets which are unobservable. We define them only for clear explanations.

To approximate $\hat{h}$, the LNGT model determines a hypothesis space $\mathcal{H}$ of hypotheses $h(\cdot;\theta)$'s, where $\theta$ denotes all the parameters used by $h$. A LNGT algorithm is an optimization strategy that searches $\mathcal{H}$ to find the $\theta$ that parameterizes the best $h^{*}\in \mathcal{H}$. The LNGT performance is measured by a loss function $\ell(\cdot,\cdot)$ defined over the prediction $h(\bm{x};\theta)$ and the observed output $y$ over the test set.

\subsection{Empirical Risk Minimization with Error Decomposition} Given a hypothesis $h$, we want to minimize its \emph{expected risk} $R$, which is the loss measured with respect to $p(\bm{x},y)$. Specifically,
\begin{align}
	\label{eq:expected_risk}
	R(h)= \int \ell(h(\bm{x}),y)dp(\bm{x},y) = \mathbb{E}[\ell(h(\bm{x}),y)].
\end{align}
As $p(\bm{x},y)$ is unknown, similar to regular machine learning tasks, the \emph{empirical risk}, i.e., the average of sample losses over the noisy training set $\bar{D}_{\text{train}}$ of $N$ samples,
\begin{align}
	\label{eq:empirical_risk}
	R_{N}(h)&=\frac{1}{|\bar{D}_{\text{train}}|}\sum_{(\bm{x},\bar{y}) \in \bar{D}_{\text{train}}}\ell(h(\bm{x}),\bar{y}) \nonumber \\ &=\underbrace{\frac{1}{|\bar{D}_{\text{c\_train}}|}\sum_{(\bm{x},\bar{y}) \in \bar{D}_{\text{c\_train}}}\ell(h(\bm{x}),\bar{y})}_{R_{N_{\text{c}}}(h)} + \underbrace{\frac{1}{|\bar{D}_{\text{m\_train}}|}\sum_{(\bm{x},\bar{y}) \in \bar{D}_{\text{m\_train}}}\ell(h(\bm{x}),\bar{y})}_{R_{N_{\text{m}}}(h)} 
\end{align}
is usually used as a proxy for $R(h)$, leading to \emph{empirical risk minimization} \citep{mohri2018foundations}. However, in this case, minimizing $R_{N}(h)$ usually leads to an estimation of $\bar{p}(\bm{x},\bar{y})$, which is completely different from $p(\bm{x},y)$. Therefore, directly training models without any adjustment has been observed to lead to poor generalization performance \citep{zhang2018understanding}. Here we can decouple the $R_{N}(h)$ into $R_{N_{\text{c}}}(h)$ and $R_{N_{\text{m}}}(h)$. Since $\bar{D}_{\text{c\_train}}$ is a set of clean samples, finding a hypothesis that only minimizes $R_{N_{\text{c}}}(h)$ rather than $R_{N_{\text{m}}}(h)$ leads to a better estimation of $p(\bm{x},y)$. For better illustration, let 
\begin{itemize}
	\item $\hat{h}=\arg \min_{h}R(h)$ be the function that minimizes the expected risk;
	\item $h^{*}=\arg \min_{h\in \mathcal{H}}R(h)$ be the function in $\mathcal{H}$ that minimizes the expected risk;
	\item $h_{N}=\arg \min_{h\in \mathcal{H}}R_{N}(h)$ be the function in $\mathcal{H}$ that minimizes the empirical risk;
	\item $h_{N_{\text{c}}}=\arg \min_{h\in \mathcal{H}}R_{N_{\text{c}}}(h)$ be the function in $\mathcal{H}$ that only minimizes the empirical risk of $\bar{D}_{\text{c\_train}}$ rather than $\bar{D}_{\text{m\_train}}$ (Assume it is achievable state during learning). 
\end{itemize}

\begin{figure*}
	\centering
	\subfigure[Learning with clean data]{
		\includegraphics[width=0.45\columnwidth]{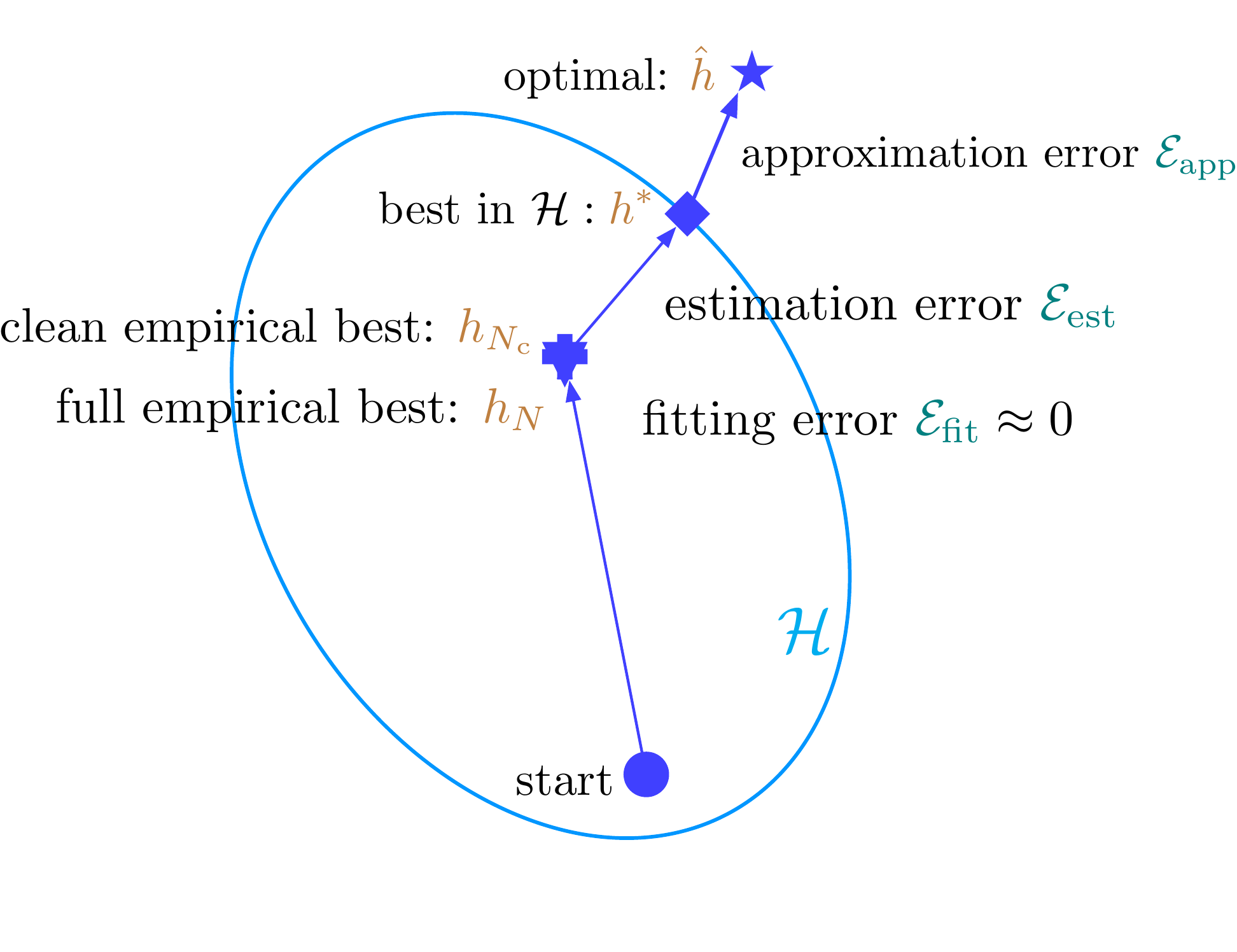}}
	\subfigure[Learning with noisy data]{
		\includegraphics[width=0.5\columnwidth]{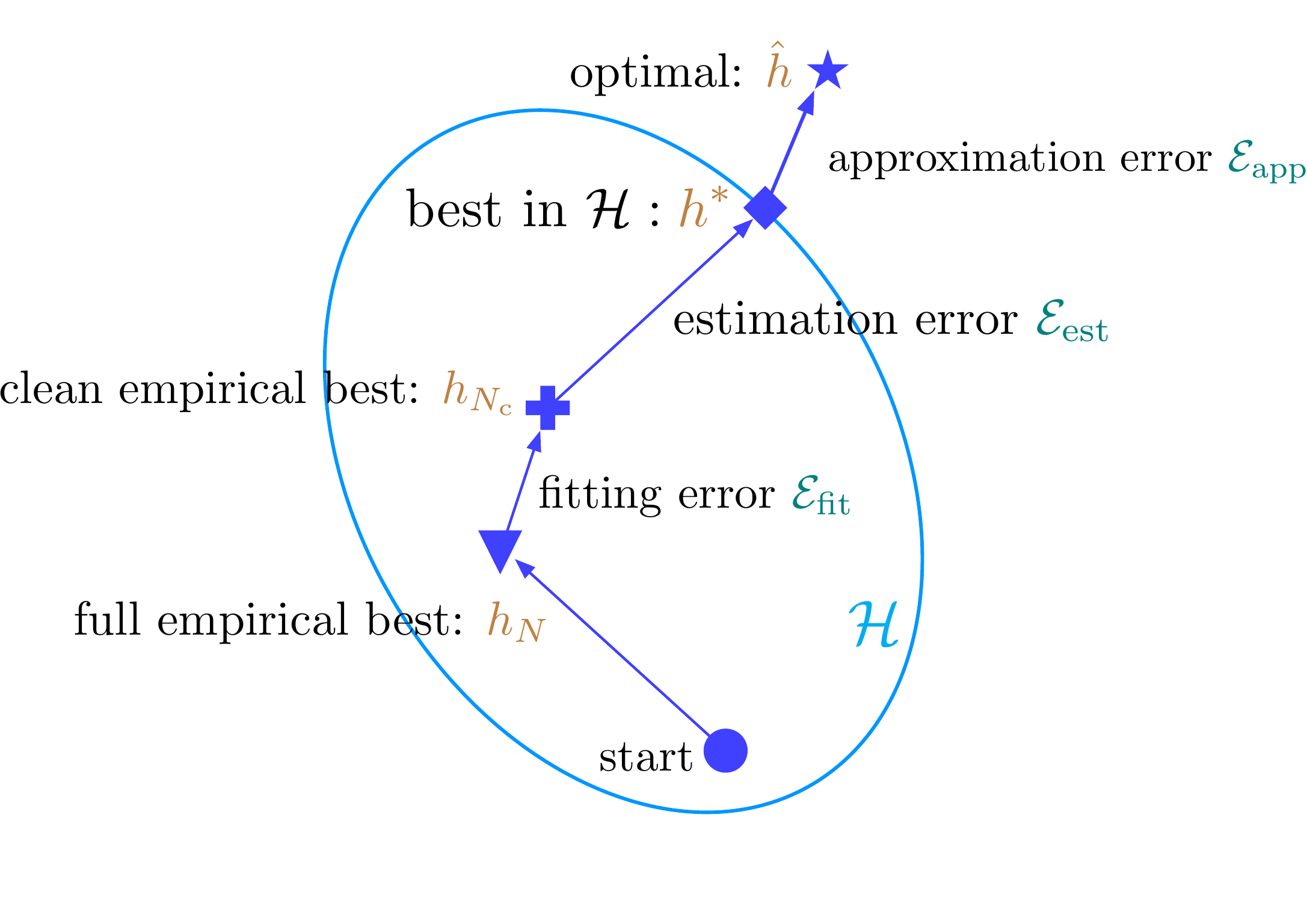}}
	\caption{Comparison of learning with clean and noisy ground truth.}
	\label{fig:intro_clean_noise}
\end{figure*}

As $\hat{h}$ is unknown, one has to approximate it by searching some $h\in \mathcal{H}$. $h^{*}$ is the best approximation for $\hat{h}$ in $\mathcal{H}$. $h_{N}$ is the best hypothesis in $\mathcal{H}$ obtained by minimizing the whole empirical risk $R_{N}(h)$, while $h_{N_{\text{c}}}$ is the optimal hypothesis in $\mathcal{H}$ that only minimizes $R_{N_{\text{c}}}(h)$. For simplicity, we assume that $\hat{h}$, $h^{*}$, $h_{N}$, and $h_{N_{\text{c}}}$ are unique. The \emph{total error} can be decomposed as
\begin{align}
	\mathbb{E}[R(h_{N})-R(\hat{h})]=\underbrace{\mathbb{E}[R(h^{*})-R(\hat{h})]}_{\mathcal{E}_{\text{app}}(\mathcal{H})} + \underbrace{\mathbb{E}[R(h_{N_{\text{c}}})-R(h^{*})]}_{\mathcal{E}_{\text{est}}(\mathcal{H},N_{\text{c}})} + \underbrace{\mathbb{E}[R(h_{N})-R(h_{N_{\text{c}}})]}_{\mathcal{E}_{\text{fit}}(\mathcal{H},N,N_{\text{c}})}, \nonumber
\end{align}
where the expectation is with respect to the random choice of $\bar{D}_{\text{train}}$. The \emph{approximation error} $\mathcal{E}_{\text{app}}(\mathcal{H})$ measures how close the functions in $\mathcal{H}$ can approximate the optimal hypothesis $\hat{h}$. The \emph{estimation error} $\mathcal{E}_{\text{est}}(\mathcal{H},N_{\text{c}})$ measures the effect of minimizing the clean empirical risk $R_{N_{\text{c}}}(h)$ instead of the expected risk $R(h)$ within $\mathcal{H}$. The \emph{fitting error} $\mathcal{E}_{\text{fit}}(\mathcal{H},N,N_{\text{c}})$ measures the effect of minimizing the full empirical risk $R_{N}(h)$ instead of only the clean empirical risk $R_{N_{\text{c}}}(h)$.

\subsection{Unreliable Empirical Risk Minimizer}
As can be observed, the total error is influenced by $\mathcal{H}$ (hypothesis space), $N$ (number of samples in $\bar{D}_{\text{train}}$) and $N_{\text{c}}$ (number of samples in $\bar{D}_{\text{c\_train}}$). A special case is when $N=N_{\text{c}}$, the LNGT reduces to regular learning problem.

Therefore, reducing the total error can be attempted from the perspectives of (1) data, which provides $\bar{D}_{\text{train}}$ and $\bar{D}_{\text{c\_train}}$; (2) model, which determines $\mathcal{H}$; and (3) algorithms, which searches for the optimal hypothesis $h_{N_{\text{c}}}$ that only fits $\bar{D}_{\text{c\_train}}$.

In LNGT, the model would easily fit all noisy samples. The empirical risk $R_{N}(h)$ may then be far from being a good approximation of the expected risk $R(h)$, and the resultant empirical risk minimizer $h_{\text{N}}$ overfits. Indeed, this is the core issue of LNGT, i.e., the empirical risk minimizer $h_{\text{N}}$ is no longer reliable. Therefore, LNGT is much harder. A comparison of learning with clean data and noisy data is shown in Figure \ref{fig:intro_clean_noise}. Compared to learning with clean data, both the estimation error and fitting error of LNGT increase.

\section{Solutions}

\begin{figure*}
	\centering
	\subfigure[Reduce $\mathcal{E}_{\text{est}}$]{
		\includegraphics[width=0.3\columnwidth]{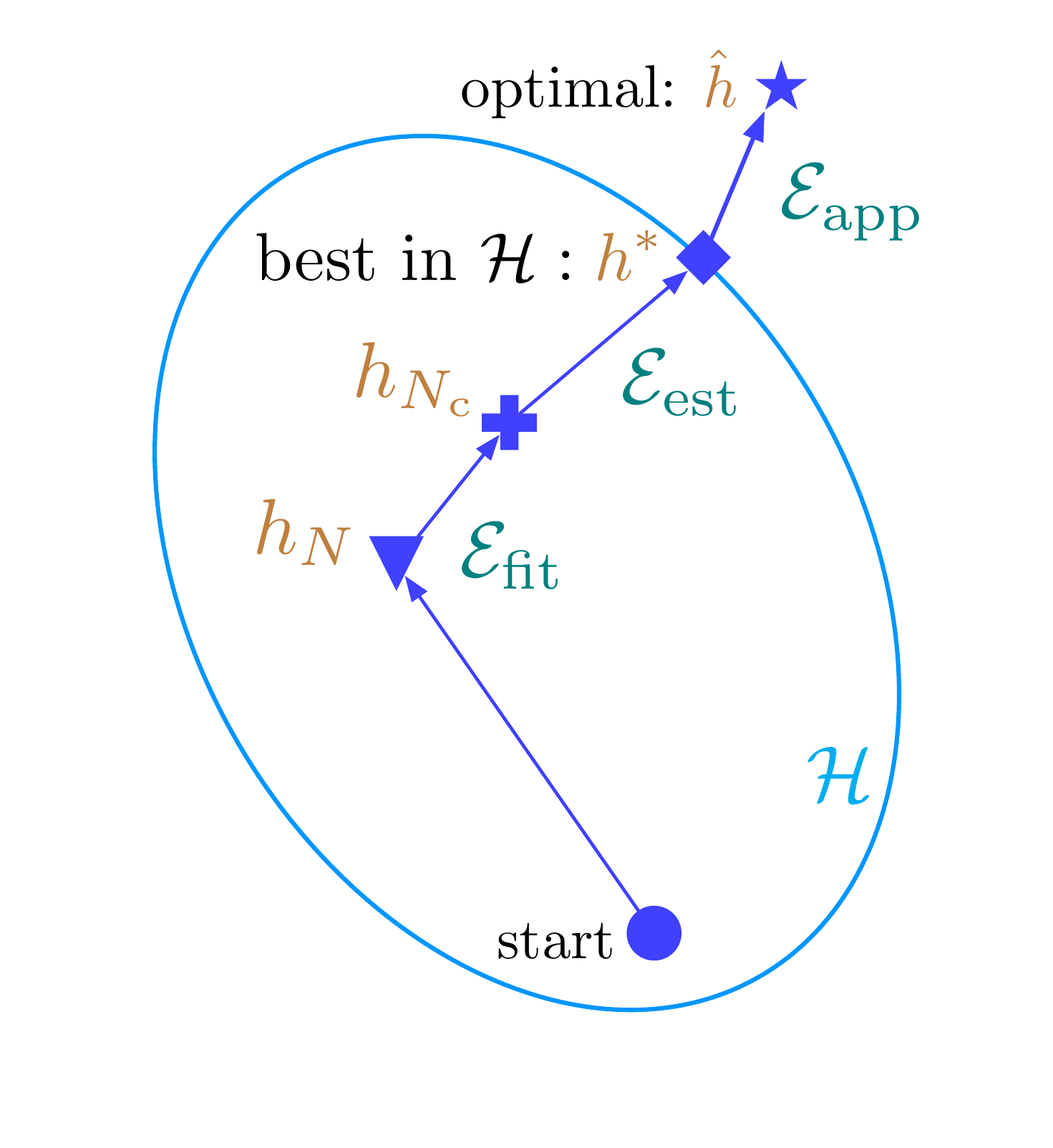}}
	\subfigure[Reduce $\mathcal{E}_{\text{fit}}$]{
		\includegraphics[width=0.3\columnwidth]{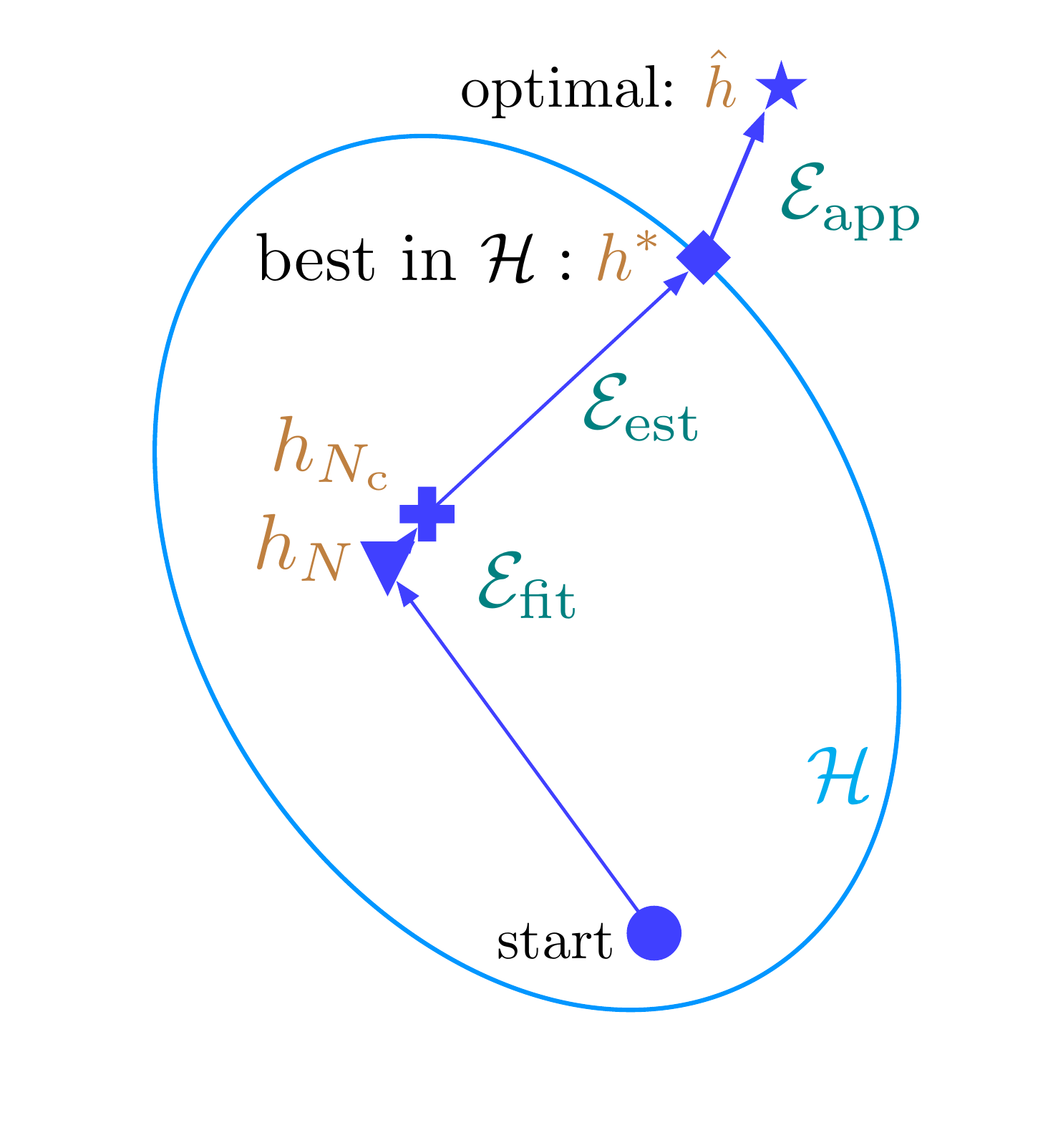}}
	\subfigure[Reduce both $\mathcal{E}_{\text{est}}$ and $\mathcal{E}_{\text{fit}}$]{
		\includegraphics[width=0.3\columnwidth]{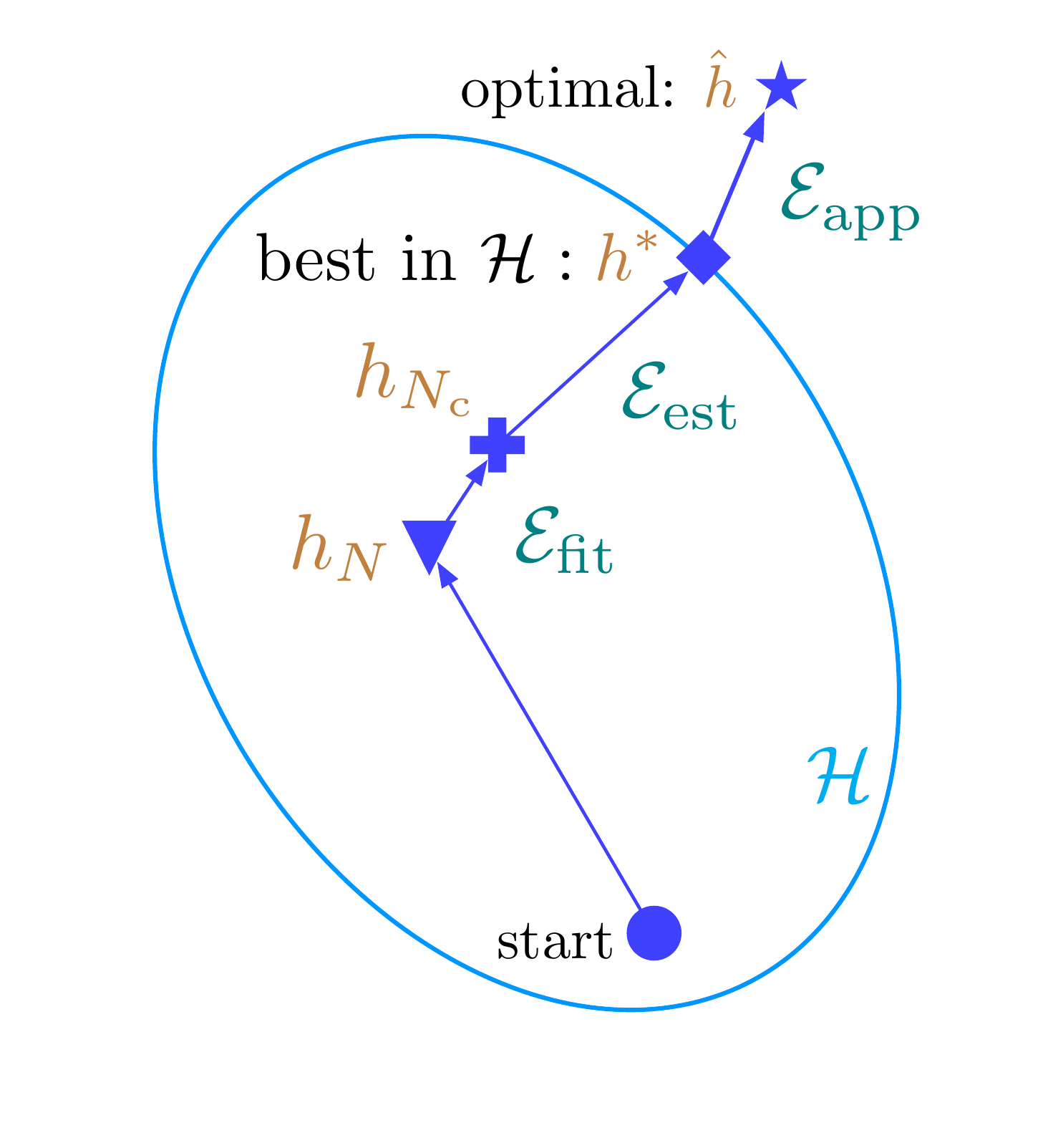}}
	\caption{Different perspectives on how existing methods solve the LNGT problem.}
	\label{fig:intro_dif_ways}
\end{figure*}

Almost all the existing works aim to reduce the learning errors to achieve noise robustness, we summarize them into three different categories (see Fig. \ref{fig:intro_dif_ways}).

\subsection{Reduce estimation error $\mathcal{E}_{\text{est}}$} 

Similar to learning with clean data, the estimation error can be reduced by increasing the number of samples. Therefore, some methods use prior knowledge to augment $\bar{D}_{\text{train}}$. For example, Mixup \citep{zhang2018mixup} constructs virtual training samples by linearly combining two random samples' features and labels. \cite{nishi2021augmentation} evaluated multiple augmentation strategies and found that using one set of augmentations for loss modeling tasks and another set for learning is the most effective in LNGT. MixNN \citep{lu2021mixnn} dynamically mixes the sample with its nearest neighbors to generate synthetic samples for noise robustness. Other methods leverage the unlabeled data to improve the performance of LNL. For instance, \cite{garg2021ratt} augmented the training data with random labeled data and provided a theoretical analysis that ensures the true risk is lower. \cite{iscen2022learning} utilized the unlabeled data to enforce the consistency of model predictions, resulting in improving the performance. Combined with Curriculum Learning, more complex Mixup-based methods \citep{li2020dividemix,cordeiro2023longremix,nagarajan2024bayesian} have been proposed recently.

\subsection{Reduce fitting error $\mathcal{E}_{\text{fit}}$}

These methods aim to prevent the model from overfitting to mislabeled samples. 

\subsubsection{Regularization}: these methods implicitly restrict the model parameters or adjusts the gradients to prevent the model from memorizing mislabeled samples. For example, \cite{li2020gradient} proved the gradient descent with early stopping is an effective regularization to achieve robustness to label noise. \cite{hu2019simple} added the regularizer to limit the distance between the model parameters to initialization for noise robustness. ELR \citep{liu2020early} estimates the target by temporal ensembling \citep{laine2016temporal} and adds a regularization term to cross-entropy loss to avoid memorization of mislabeled samples. NAL \citep{lu2022noise} scales the gradients according to the cleanliness of different samples to achieve noise robustness. 

\subsubsection{Robust loss functions}: these methods develop loss functions that are inherently robust to label noise, including DMI \citep{xu2019l_dmi}, MAE \citep{ghosh2017robust}, GCE \citep{zhang2018generalized}, SCE \citep{wang2019symmetric}, NCE \citep{ma2020normalized}, TCE \citep{feng2021can}, GJS \citep{englesson2021generalized} and CE+EM \citep{lu2022robust}. These methods are to hypothesize noise models and then develop robust algorithms based on them. Two typical noise assumptions are symmetric and asymmetric label noise \cite{natarajan2013learning,patrini2017making}, where the labels are corrupted by a $K\times K$ noise transition matrix $Q$ (where $K$ is the number of class in classification task), i.e., $Q_{ij}=P(\bar{y}=j|y=i)$, where $y$ denotes the true label and $\bar{y}$ denotes the noisy label. Suppose noise rate is $\eta$, for symmetric noise, the flip probability to other labels is constant, i.e., $Q_{ij}=1-\eta \text{ for }i=j$ and $Q_{ij}=\frac{\eta}{K-1}$ for $i\ne j$. For asymmetric noise, it is a simulation of real-world label noise, where labels are only replaced by similar classes.

\begin{table*}[t]
	\centering
	\resizebox{1\textwidth}{!}{
		\begin{tabular}{c|c|c|c|c}
		Methods & Loss expression &$\sum_{i=1}^{K}\ell(f(\bm{x}),i)$& Symmetric & Gradient $\frac{ \partial \ell (f(x),y)}{\partial \theta}$ \\ 
		\hline
		CE & $-\log p(y|\bm{x})$& $-\sum_{i=1}^{K}\log p(i|\bm{x})$& $\times$ & $-\frac{1}{p(y|\bm{x})}\nabla_{\theta}p(y|\bm{x})$ \\
		FL  & $-(1-p(y|\bm{x}))^{\gamma}\log p(y|\bm{x})$ & $-\sum_{i=1}^{K}(1-p(i|\bm{x}))^{\gamma}\log p(i|\bm{x})$ & $\times$ & $\Big[\gamma(1-p(y|\bm{x}))^{\gamma-1}\log p(y|\bm{x})-\frac{(1-p(y|\bm{x}))^\gamma}{p(y\bm{x})}\Big]\nabla_{\theta}p(y|\bm{x})$ \\
		MAE  & $2(1-p(y|\bm{x}))$& $2K-2$& $\surd$ & $-2\nabla_{\theta}p(y|\bm{x})$ \\
		RCE  &$-A(1-p(y|\bm{x}))$ & $-AK+A$& $\surd$ & $A\nabla_{\theta}p(y|\bm{x})$ \\
		GCE  & $\frac{1-p(y|\bm{x})^{\rho}}{\rho}$ & $[\frac{K-K^{1-\rho}}{\rho},\frac{K-1}{\rho}]$& $\surd$ & $-\frac{1}{p(y|\bm{x})^{1-\rho}}\nabla _{\theta}p(y|\bm{x})$ \\
		TCE   &$\sum_{i=1}^{t}\frac{(1-p(y|\bm{x}))^{i}}{i}$ & $\big[K-1,(K-1)\sum_{i=1}^{t}\frac{1}{i}\big]$ & $\surd$ & $-\frac{1-(1-p(y|\bm{x}))^{t}}{p(y|\bm{x})}\nabla_{\theta}p(y|\bm{x})$\\
		NCE  & $\log_{\Pi^{K}_{k}p(k|\bm{x})}p(y|\bm{x})$ & 1& $\surd$ & $\frac{\sum_{k\ne y}^{K}\log p(k|\bm{x})}{(\sum_{k=1}^{K}\log p(k|\bm{x}))^{2}}\cdot\frac{1}{p(y|\bm{x})}\nabla _{\theta}p(y|\bm{x})$ \\
	\end{tabular}
}
\caption{Comparison of existing loss functions. For FL, parameter $\gamma\ge 0$ and FL reduces to the CE loss when $\gamma=0$. For RCE, $A$ is a negative constant to replace $\log(0)$. For GCE, parameter $\rho\in(0,1]$. For TCE, parameter $t \in \mathbb{N}_{+}$.}  
\label{tab:losses}
\end{table*}

Given symmetric noise rate $\eta < \frac{K-1}{K}$, a loss function is proved to be noise-tolerant if it satisfies the symmetric condition as follows \cite{ghosh2017robust}:
\begin{align}
	\sum_{k=1}^{K}\ell(f(\bm{x}),k)=C, \forall \bm{x}\in \mathcal{X}, \forall f \in \mathcal{H},
\end{align}
where $C$ is a constant, and $\mathcal{H}$ is the hypothesis class. Then we can easily obtain 
\begin{align}
	R_{\ell}^{\eta}(f)=\big(1-\frac{\eta K}{K-1}\big)R_{\ell}(f) + \frac{\eta C}{K-1}.
\end{align}
Since $1-\frac{\eta K}{K-1}>0$, if $f^{*}$ is the global minimizer of $R_{\ell}(f)$, then it is also the minimizer of $R_{\ell}^{\eta}(f)$. Therefore, a symmetric loss function is theoretical noise-tolerant if the global minimizer can be learned. However, the derivation of global optimum is a strong assumption. 
In practice, many robust loss functions have been observed to suffer from the underfitting problem on complicated datasets \cite{zhang2018generalized,ma2020normalized}. We review the existing loss functions and derive their gradients in Table \ref{tab:losses}. The CE loss and focal loss (FL) \cite{lin2017focal} are not robust to noisy labels but have the advantage of sufficient learning ability. Both of them put more weights on the gradient of ambiguous (hard) samples. On the contrary, MAE and Reverse CE (RCE) \cite{wang2019symmetric} are robust to noisy labels but increase difficulty in training as they equally provide the same weights on the gradient for all training samples. To balance learning sufficiency and noise robustness, a generalized version of CE loss (GCE) \cite{zhang2018generalized} was proposed $\ell_\text{gce}=\frac{1-p(y|\bm{x})^{\rho}}{\rho}$, which reduces to MAE and CE when $\rho=1$ and $\rho\rightarrow0$, respectively. Similarly, Taylor cross entropy (TCE) \cite{feng2021can} loss was proposed $\ell_\text{tce}=\sum_{i=1}^{t}\frac{(1-p(y|\bm{x}))^{i}}{i}$, which is also a generalized mixture of CE (when $t\rightarrow \infty$) and MAE (when $t=1$). Recently, \citet{ma2020normalized} have demonstrated that any loss can be made robust to noisy labels by applying a simple normalization, e.g., normalized cross entropy (NCE). However, the normalization operation actually alters the gradient of CE loss so that NCE  no longer retains the original fitting ability. Let's denote $P=\log p(y|\bm{x})$ and $Q=\sum_{k\ne y}\log p(k|\bm{x})$. In Table \ref{tab:losses}, the gradient of NCE is weighted by the term $\frac{\sum_{k\ne y}^{K}\log p(k|\bm{x})}{(\sum_{k=1}^{K}\log p(k|\bm{x}))^{2}}=\frac{Q}{(P+Q)^{2}}$. During training, the $Q$ term may increase even when $P$ is fixed. $Q$ reaches the maximum value when all $p(k\ne y|\bm{x})$ equals to $(1-p(y|\bm{x}))/(K-1)$. As a consequence, the corresponding gradient reaches the minimum value, which hinders the convergence and causes the underfitting problem. To solve this problem, Active Passive Loss (APL) \cite{ma2020normalized} was proposed for both robust and sufficient learning by combining two loss terms.

\subsubsection{Sample selection}: The key idea is trying to select clean samples or reweigh the samples in training. During the early learning stage, the samples with smaller loss values are more likely to be the clean samples. Based on this observation, MentorNet \citep{jiang2018mentornet} pre-trains a mentor network for assigning weights to samples for guiding the training of the student network. Decoupling \citep{malach2017decoupling} updates the two networks by using the samples having different predictions. Co-teaching \citep{han2018co} trains two networks which select small-loss samples within each mini-batch to train each other. Co-teaching+ \citep{yu2019does} improves it by updating the network on disagreement data to keep the two networks diverged. \cite{ren2018learning} reweighed samples based on their gradient directions. JoCoR \citep{wei2020combating} jointly trains two networks with the examples that have prediction agreement between two networks. Co-matching \citep{lu2022ensemble} uses a novel framework with two networks fed with different strengths of augmented inputs to achieve the better ensemble effect.

\subsubsection{Loss correction}: These methods correct the loss by estimating the noise transition matrix. \cite{patrini2017making} estimated the label corruption matrix for loss correction. \cite{hendrycks2018using} improved the corruption matrix by using a clean set of data.

\subsection{Reduce both $\mathcal{E}_{\text{est}}$ and $\mathcal{E}_{\text{fit}}$}

Some methods focus on correcting the noisy labels, so that the model is gradually refined. \cite{reed2015training} proposed a bootstrapping method which modifies the loss with model predictions. \cite{ma2018dimensionality} improved the bootstrapping method by exploiting the dimensionality of feature subspaces to dynamically reweigh the samples. \cite{pmlr-v97-arazo19a} improved bootstrapping using a dynamic weighting scheme through unsupervised learning techniques. PLC \citep{zhang2020learning} progressively corrects the labels when the prediction confidence over a dynamic threshold. SELC \citep{lu2022selc} gradually corrects noisy labels by ensemble predictions.


\bibliographystyle{ACM-Reference-Format}
\bibliography{sample-base}

\appendix

\end{document}